\documentclass[10pt,twocolumn,letterpaper]{article}

\usepackage{wacv}              %

\definecolor{wacvblue}{rgb}{0.21,0.49,0.74}
\usepackage[pagebackref,breaklinks,colorlinks,allcolors=wacvblue]{hyperref}
\usepackage[margin=0.75in]{geometry}
\usepackage{times}
\usepackage{amsmath}
\usepackage{amssymb}
\usepackage{amsthm}
\usepackage{algorithm}
\usepackage{algpseudocode}
\usepackage{booktabs}
\usepackage{capt-of}
\usepackage{xcolor}
\usepackage{multirow}
\definecolor{Darkgreen}{RGB}{1,100,32}

\newcommand{\method}{\textsc{RecCAR}}
\def\wacvPaperID{*****} %
\def\confName{WACV}
\def\confYear{2027}

\newcommand\ignore[1]{}

\newcommand\ourmethodlong{Reciprocal Cross-modal Attention Regularization}
\newcommand\ourmethod{RecCAR}

\newcommand\dvir[1]{\textcolor{Darkgreen}{[DS: #1]}}

\title{All modalities are equal, but video is more equal: \\ Closing the Cross-Attention Gap in Joint Video Generation}

\author{
Ohad Rahamim$^{1}$ \quad
Dvir Samuel$^{2}$ \quad
Idan Schwartz$^{1}$ \quad
Gal Chechik$^{1,2}$\\
$^{1}$Bar-Ilan University \quad
$^{2}$NVIDIA
}

\makeatletter
\let\@oldmaketitle\@maketitle
\renewcommand{\@maketitle}{\@oldmaketitle
  \begin{center}
    \centering
    \includegraphics[width=\textwidth]{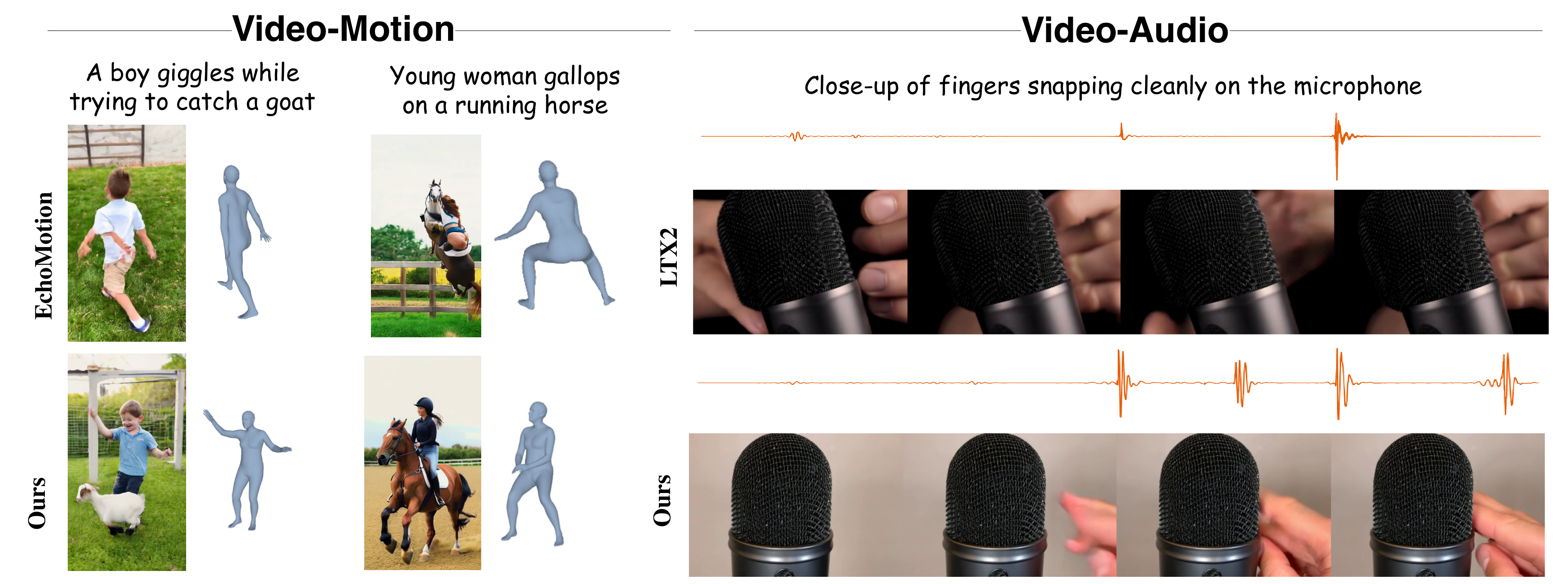}
    \captionof{figure}{
Overview of our method, \method{}, compared to  EchoMotion \cite{yang2026echomotion} and LTX-2 \cite{hacohen2026ltx2}, across three examples. \textbf{Left:} For Video-Motion, we show two examples of a generated video frame alongside its SMPL motion. Top: baseline results with inaccurate or physically implausible anatomy. Bottom (+ \ourmethod{}): improved video--motion alignment. \textbf{Right:} For Audio--Video, we show the audio waveform and corresponding video frames. Top: the video contains no tapping sounds. Bottom (+ \ourmethod{}): the tapping action is correctly synchronized between audio and video.
    }
    \label{fig:teaser}
  \end{center}
  \vspace{15pt}
}
\makeatother

\begin{document}
\maketitle
\begin{abstract}
    Video is a rich representation of a physical event, capturing appearance, geometry, motion, and temporal evolution. Other modalities, such as 3D body motion or audio, encode narrower aspects of the same event. We find that joint multimodal diffusion transformers exhibit a corresponding asymmetry in cross-modal correspondence: companion modalities develop strong correspondences to video, but the reciprocal correspondences through which they constrain video remain substantially weaker. We express both directions as comparable correspondence distributions over video tokens and define their disagreement as the reciprocal correspondence gap. We introduce \method{}, standing for \emph{Reciprocal Cross-modal Attention Regularization}, a KL regularizer that uses the well-established video-to-modality correspondence as a fixed reference and aligns the weaker modality-to-video correspondence toward it. Across joint video--motion and video--audio generation, \method{} improves the Human Anatomy score from 0.69 to 0.75 and reduces audio--video desynchronization from 0.804 to 0.752, while improving overall generation quality. project page: \href{https://ohad204.github.io/RecCAR.github.io/}{https://ohad204.github.io/RecCAR.github.io/}
\end{abstract}

\section{Introduction}
\label{sec:intro}

Modern generative models can produce multi-modal content in which all modalities are generated jointly rather than in isolation. This is the case, for example, with generation of video together with audio~\cite{ruan2023mmdiffusion,liu2026javisdit,low2025ovi,zhang2025uniavgen,hacohen2026ltx2,omniforcing,zhao2025uniform,sun2025proav,team2026mova,wang2025universe,hu2026harmony,ma2026improving,chen2026just,huang2025jova,chen2026skyreels}, with human  motion~\cite{chefer2025videojam,yang2026echomotion,zhao2026comovi,wang2025vimogen,wang2025mosa}, with flow~\cite{chefer2025videojam}, or with depth~\cite{kwon2025jointdit,zhai2024idol}.
Unlike cascaded pipelines, where one output is generated first and the second is predicted afterward, joint generation allows all modalities to influence one another throughout the denoising process, yielding outputs that are not only individually realistic but also mutually consistent.

Most recent joint diffusion models implement cross-modal interaction through bidirectional cross-modal attention~\cite{hacohen2026ltx2,yang2026echomotion}.
Each modality can attend to the representation of the other, establishing two reciprocal directions of cross-modal correspondence. Architecturally, the two generation streams are therefore connected in both directions.
The standard generative objective, however, places no direct constraint on the agreement between reciprocal cross-modal correspondences. Consequently, the two attention directions can develop substantially different correspondence patterns, even when both are supported by the model structure.

We find that this asymmetry is pronounced in pretrained joint multimodal generators.
Although both attention directions are available, in practice one modality often develops substantially stronger and more informative cross-modal correspondences than the other.
The result is models that are \emph{architecturally bidirectional but functionally asymmetric}: one modality adapts to the other, while the reciprocal influence remains largely inactive.
This weakens one of the main motivations for joint generation, since information available in one generated stream may fail to correct inconsistencies in the other.

Importantly, the stronger attention direction already contains useful supervision for the weaker one.
Both directions describe interactions between the same pair of modalities, only viewed from opposite sides.
When one direction has learned a meaningful cross-modal correspondence, that correspondence can serve as an internal target for its reciprocal pathway. The model can therefore improve reciprocal correspondence alignment using structure that is already present in its own pretrained representations, without requiring an external leader or additional annotations.

Based on this observation, we introduce \textit{\ourmethodlong{}} (\ourmethod{}), a lightweight training strategy for aligning reciprocal cross-modal correspondences in joint multimodal generators.
\ourmethod{} treats the stronger cross-attention direction as a fixed leader and aligns the reciprocal direction to it through a stop-gradient KL objective.
In this way, training transfers cross-modal structure from the better-established pathway to the weaker one, so both can be used later during inference.
The approach operates directly on the model's existing cross-attention maps and requires no additional inference-time component. We fine-tune only lightweight LoRA parameters, preserving the capabilities of the original pretrained generator.

A key property of \ourmethod{} is that it is agnostic to the particular modality being generated with the video (the companion stream). We demonstrate its efficacy in two substantially different joint-generation settings. First, in \textbf{video--human-motion generation}, the companion stream consists of structured 3D body motion.Applied to EchoMotion~\cite{yang2026echomotion}, \ourmethod{} reduces N-MPJPE from $0.197$ to $0.148$ and improves the VBench Human Anatomy score from $0.69$ to $0.75$ while preserving motion dynamics and visual quality. Standard fine-tuning on the same data does not obtain these gains.

As a second task, we evaluate \ourmethod{} on \textbf{video--audio generation}, where the companion modality carries very different temporal and semantic information.
When applied to LTX-2~\cite{hacohen2026ltx2}, \ourmethod{} reduces absolute audio--video desynchronization from $0.804$ to $0.752$ on T2AV-Compass~\cite{cao2026t2avcompass} and further improves synchronization on AVGen-Bench~\cite{zhou2026avgen}, while preserving audio quality, video quality, and semantic alignment.
The same reciprocal-attention objective therefore improves cross-modal consistency across both spatially structured motion and temporally structured audio.

Together, these results highlight a broader limitation of current joint multimodal generators:
\emph{bidirectional connectivity does not guarantee aligned bidirectional correspondences}.
Simply coupling two generative streams is insufficient when one modality learns to dominate their interaction.
\ourmethod{} provides a simple mechanism for closing this gap by transferring cross-modal knowledge from the stronger attention pathway to its reciprocal counterpart.

\paragraph{Contributions.}
Our main contributions are:
(1) we identify a systematic asymmetry between reciprocal cross-attention directions in joint multimodal generators, showing that architecturally bidirectional models can exhibit substantially different correspondence patterns in their two attention directions;
(2) we introduce \ourmethod{}, a lightweight reciprocal cross-attention regularization objective that directly aligns the weaker cross-modal correspondence map with its well-established reciprocal counterpart; and
(3) we demonstrate the generality of this principle across two substantially different settings: video--motion and video--audio generation, improving cross-modal consistency while preserving the quality of the underlying generators.

\section{Related Works}

\subsection{Modality Imbalance}
\label{sec:related-imbalance}

Multimodal models do not necessarily use all available modalities equally.
Prior work has shown that different modalities may be learned at different
rates~\cite{wang2020makes}, allowing one modality to dominate the other
\cite{peng2022balanced,fan2023pmr}. Gat et al.~\cite{gat2020removing}
similarly identified strong modality preferences in multimodal classifiers and
proposed regularization to reduce them, while Perceptual
Score~\cite{gat2021perceptual} quantified which modalities a trained model
actually relies on. We study a related asymmetry within joint generation:
bidirectional cross-attention provides reciprocal pathways, but the
video-to-modality correspondence can be substantially better established than
its modality-to-video counterpart.

\subsection{Attention Optimization}
Cross-attention is widely used to connect modalities and condition diffusion
models~\cite{hertz2022prompt,zhang2023controlnet,ye2023ipadapter,liu2023zero123}.
Its internal maps also provide useful optimization targets:
Prompt-to-Prompt~\cite{hertz2022prompt} showed that diffusion cross-attention
encodes meaningful spatial correspondences, Chefer et
al.~\cite{chefer2022optimizing} directly optimized attention-derived relevance
maps to improve robustness, and related work manipulated attention maps to
improve spatial or semantic control
\cite{voleti2024zerotohero,chefer2023attend}. 

\subsection{Joint Multi-Modal Diffusion}
\label{sec:related-joint}

Joint diffusion models generate multiple modalities within the same denoising
process, allowing them to interact rather than treating one as a fixed
condition. This paradigm has been explored for RGB-depth generation
\cite{kwon2025jointdit,zhai2024idol,liu2025idcnet}, and more extensively for
joint audio-video and video-motion generation.

Audio-visual generation evolved from directional approaches that adapted
visual generators to audio~\cite{yariv2023audiotoken,yariv2024diverse} or
generated audio from fixed video
\cite{iashin2021taming,sheffer2023hear,luo2023difffoley,
zhang2024foleycrafter,cheng2025mmaudio}. MM-Diffusion
\cite{ruan2023mmdiffusion} introduced joint audio-video denoising, followed by
diffusion transformers that increasingly couple the two generated streams
throughout generation
\cite{low2025ovi,zhang2026uniavgen,wang2025universe,hacohen2026ltx2,
team2026mova,liu2026javisdit,hu2026harmony,ma2026improving}.

Similarly, early video-motion methods used motion primarily as a condition for
video generation~\cite{nam2025cameo,wang2025mosa}, while recent approaches
couple motion and video through auxiliary motion representations, joint
denoising, cross-attention, transferred video priors, mesh tokens, or
pretrained-model guidance
\cite{chefer2025videojam,yang2026echomotion,zhao2026comovi,
wang2025vimogen,liang2026meshtoken,shaulov2025flowmo}.
These joint models enable reciprocal interaction between generated modalities,
but generally leave the correspondence learned by the two directions
unconstrained. \method{} specifically regularizes these reciprocal
correspondences, using the well-established video-to-modality correspondence
as an internal reference for the weaker pathway back into video.

\begin{figure}[t]
    \centering
    \includegraphics[
        width=1.0\linewidth,
        trim=12 8 12 8,
        clip
    ]{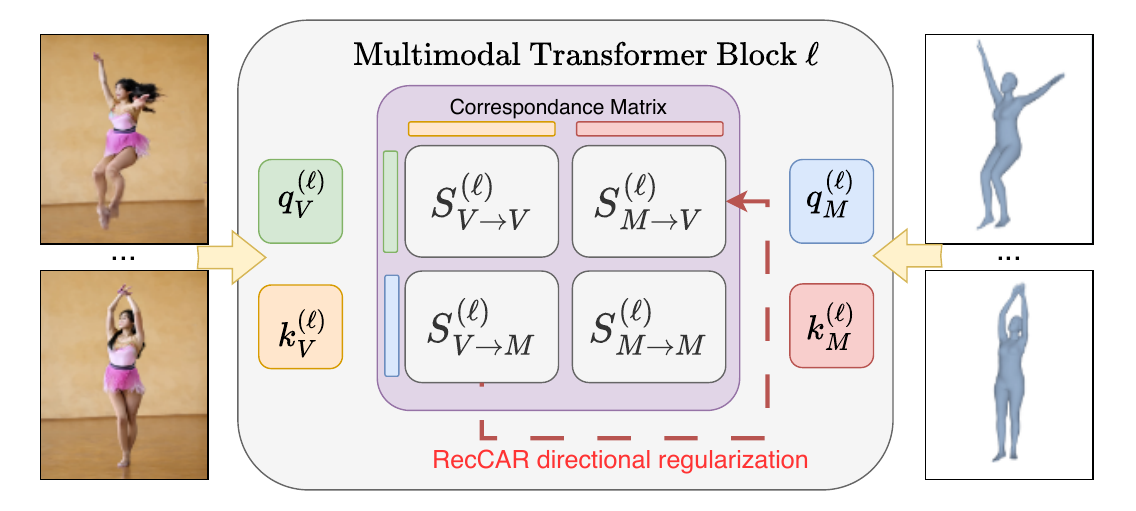}
    \caption{
    Overview of \method{} at block $\ell$. The red path indicates the directional
    \method{} regularization, which uses the video-to-modality correspondence as
    the fixed reference to strengthen the reciprocal modality-to-video pathway.
    }
    \label{fig:reccar_overview}
\end{figure}

\section{Method}
\label{sec:method}

We first describe the joint multimodal generation setting and its attention
structure. We then express the reciprocal cross-modal interactions as
comparable correspondence distributions and introduce \method{}, which uses
the better-established video-to-modality correspondence to strengthen the
reciprocal pathway back into video.

\subsection{Preliminaries: Joint Multimodal Generation}
\label{sec:preliminaries}

We consider joint generative models that produce video $V$ together with a
companion modality $M$, such as 3D human motion or audio, from the same text
prompt. Unlike cascaded approaches, both streams are generated within the
same denoising or flow-matching process and can interact throughout
generation. Figure~\ref{fig:reccar_overview} illustrates the corresponding
multimodal transformer block and the interactions between the two streams.

At transformer block $\ell$, let
$
X_V^{(\ell)}\in\mathbb{R}^{N_V\times d},
X_M^{(\ell)}\in\mathbb{R}^{N_M\times d}
$
 denote the video and companion-modality representations, where $N_V$ and
$N_M$ are their respective numbers of tokens and $d$ is the hidden
dimension. As shown in Fig.~\ref{fig:reccar_overview}, the attention block
contains intra-modal interactions, $V\!\rightarrow\!V$ and
$M\!\rightarrow\!M$, together with reciprocal cross-modal interactions,
$V\!\rightarrow\!M$ and $M\!\rightarrow\!V$. We focus on the latter, which determines how one generated stream influences the other.

Throughout, arrows denote the direction of influence:
$V\!\rightarrow\!M$ updates the modality stream using video, while
$M\!\rightarrow\!V$ updates the video stream using the companion modality.
Although the architecture permits information exchange in both directions,
the two pathways need not develop equally strong cross-modal correspondences.

\subsection{Reciprocal Cross-Modal Correspondences}
\label{sec:cross_modal_correspondences}

We express the two cross-modal directions as comparable correspondence
distributions. Using standard query and key projections, let
$q_{V,i}^{(\ell)},k_{V,i}^{(\ell)},q_{M,j}^{(\ell)},k_{M,j}^{(\ell)}
\in\mathbb{R}^{d_k}$ denote the projected video and modality tokens. For
clarity, we omit the attention-head index and backbone-specific positional
terms. The pre-softmax compatibility scores are
\begin{align}
S_{V\rightarrow M}^{(\ell)}(j,i)
&=
\frac{
\left\langle q_{M,j}^{(\ell)},k_{V,i}^{(\ell)}\right\rangle
}{\sqrt{d_k}},
\label{eq:v2m_logits}
\\
S_{M\rightarrow V}^{(\ell)}(i,j)
&=
\frac{
\left\langle q_{V,i}^{(\ell)},k_{M,j}^{(\ell)}\right\rangle
}{\sqrt{d_k}},
\label{eq:m2v_logits}
\end{align}
where $i\in\{1,\ldots,N_V\}$ and $j\in\{1,\ldots,N_M\}$ index video and
companion-modality tokens, respectively.

For $V\!\rightarrow\!M$, modality tokens query video tokens, and the native
attention already defines a distribution over video tokens:
\begin{equation}
C_{V\rightarrow M}^{(\ell)}(i\mid j)
=
\operatorname{softmax}_{i}
S_{V\rightarrow M}^{(\ell)}(j,i).
\label{eq:v2m_correspondence}
\end{equation}
Thus, for each modality token $j$,
$C_{V\rightarrow M}^{(\ell)}(\cdot\mid j)$ describes where that token
corresponds in the video.

The reciprocal $M\!\rightarrow\!V$ attention is natively normalized over
modality tokens, since each video token queries the modality stream. To make
the two directions directly comparable, we instead normalize the same logits
over video tokens:
\begin{equation}
C_{M\rightarrow V}^{(\ell)}(i\mid j)
=
\operatorname{softmax}_{i}
S_{M\rightarrow V}^{(\ell)}(i,j).
\label{eq:m2v_correspondence}
\end{equation}
Equation~\eqref{eq:m2v_correspondence} is not the native forward-pass
attention distribution, but a re-normalization of the same attention logits.
Both directions therefore answer the same question: for modality token $j$,
which video tokens are most strongly associated with it?

\begin{figure}
    \centering
    \includegraphics[width=0.9\linewidth]{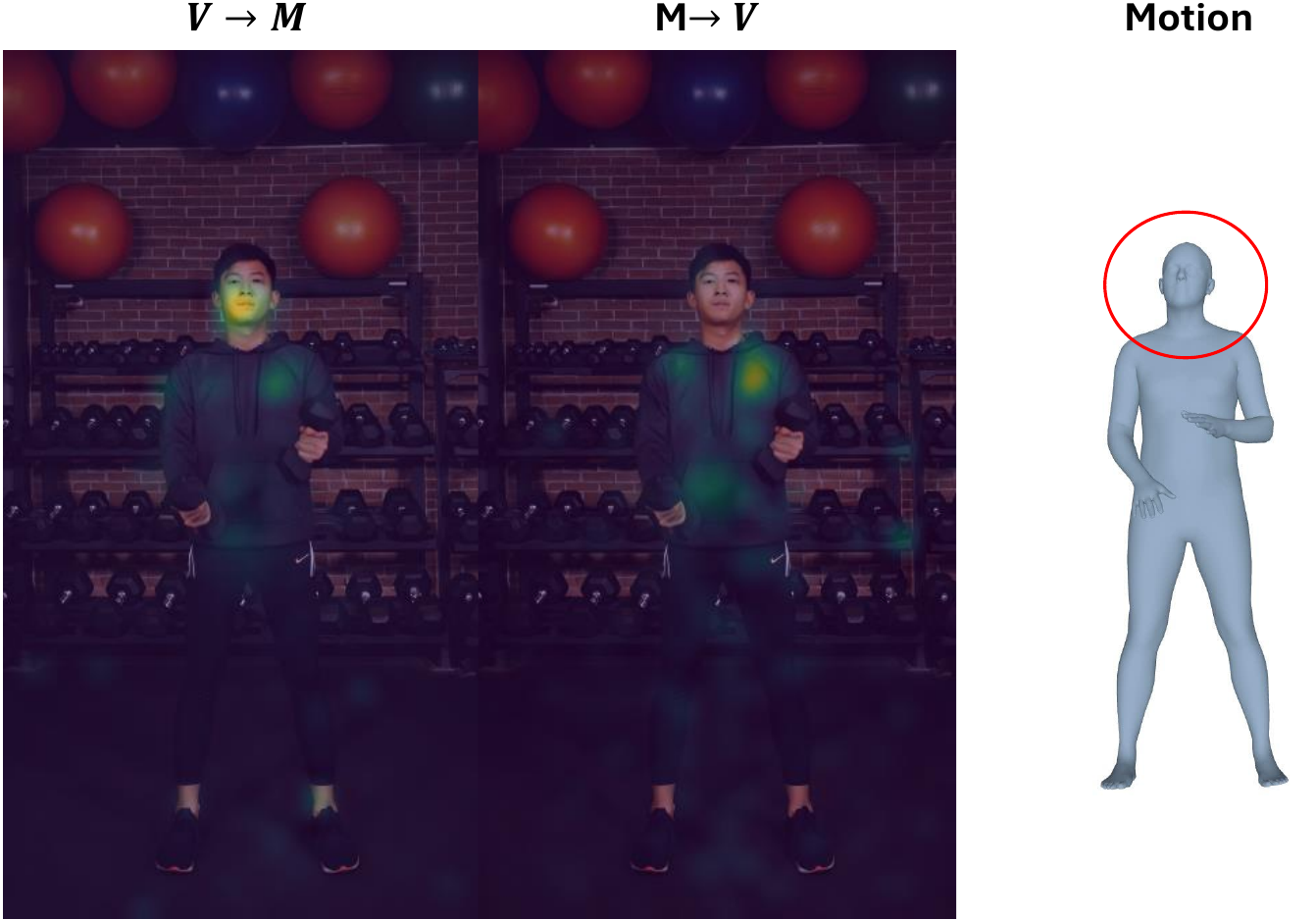}
    \caption{Illustration of asymmetric reciprocal cross-modal correspondence. For a motion token corresponding to the head (right, red circle), the \(V\!\rightarrow\!M\) correspondence correctly localizes the head region in the video, while the reciprocal \(M\!\rightarrow\!V\) correspondence is less well aligned, focusing primarily on the upper torso.}
    \label{fig:MV_vs_VM}
\end{figure}

For example, in Fig.~\ref{fig:MV_vs_VM}, if $j$ denotes the head-motion token,
$C_{V\rightarrow M}^{(\ell)}(\cdot\mid j)$ localizes the head, whereas
$C_{M\rightarrow V}^{(\ell)}(\cdot\mid j)$ places more mass on the upper
torso. Ideally, both distributions should associate the token with the same
visual region. The same interpretation extends to audio over spatiotemporal
video tokens.

\subsection{\method{}: Reciprocal Cross-modal Attention Regularization}
\label{sec:reccar}

The reciprocal correspondence distributions need not agree. In the pretrained
joint generators we study, the $V\!\rightarrow\!M$ pathway develops a
well-established correspondence, whereas the reciprocal
$M\!\rightarrow\!V$ pathway remains substantially weaker.
Figure~\ref{ig:MV_vs_VM} illustrates this asymmetry and the
directional regularization introduced by \method{}.

\paragraph{Reciprocal correspondence gap.}
Since both directions are expressed as distributions over the same video
tokens, we can directly quantify their disagreement. For a video--modality
pair $(V,M)$, we define the reciprocal correspondence gap as
\begin{equation}
\Delta_{\mathrm{corr}}^{V,M}
=
\frac{1}{|\mathcal{R}|N_M}
\sum_{\ell\in\mathcal{R}}
\sum_{j=1}^{N_M}
D_{\mathrm{KL}}\!\left(
C_{V\rightarrow M}^{(\ell)}(\cdot\mid j)
\,\middle\|\,
C_{M\rightarrow V}^{(\ell)}(\cdot\mid j)
\right),
\label{eq:correspondence_gap}
\end{equation}
where $\mathcal{R}$ denotes the considered cross-attention layers, and the
quantity is additionally averaged over attention heads. A small
$\Delta_{\mathrm{corr}}^{V,M}$ indicates similar reciprocal correspondences,
while a large value indicates disagreement between them.

\paragraph{Directional regularization.}
To close this gap, \method{} keeps the well-established
$V\!\rightarrow\!M$ correspondence fixed and optimizes the reciprocal
$M\!\rightarrow\!V$ pathway toward it. As illustrated by the red path in
Fig.~\ref{fig:reccar_overview}, we denote the fixed correspondence by
\begin{equation}
\widehat{C}_{V\rightarrow M}^{(\ell)}(i\mid j)
=
\operatorname{sg}\!\left[
C_{V\rightarrow M}^{(\ell)}(i\mid j)
\right],
\label{eq:fixed_correspondence}
\end{equation}
where $\operatorname{sg}[\cdot]$ denotes stop-gradient. The resulting
directional correspondence gap is
\begin{equation}
\widehat{\Delta}_{\mathrm{corr}}^{V,M}
=
\frac{1}{|\mathcal{R}|N_M}
\sum_{\ell\in\mathcal{R}}
\sum_{j=1}^{N_M}
D_{\mathrm{KL}}\!\left(
\widehat{C}_{V\rightarrow M}^{(\ell)}(\cdot\mid j)
\,\middle\|\,
C_{M\rightarrow V}^{(\ell)}(\cdot\mid j)
\right).
\label{eq:directional_correspondence_gap}
\end{equation}
Thus, the established video-to-modality correspondence provides a target
cross-modal map, while the weaker reciprocal pathway is encouraged to recover
the same structure. This strengthens the pathway through which the companion
modality can constrain the generated video.

\paragraph{Optimization.}
We use the directional correspondence gap as the \method{} regularizer,
$\mathcal{L}_{\mathrm{RecCAR}}
=\widehat{\Delta}_{\mathrm{corr}}^{V,M}$, and optimize
\begin{equation}
\mathcal{L}_{\mathrm{total}}
=
\mathcal{L}_{\mathrm{gen}}
+
\lambda_{\mathrm{RecCAR}}
\mathcal{L}_{\mathrm{RecCAR}},
\label{eq:total_loss}
\end{equation}
where $\mathcal{L}_{\mathrm{gen}}$ is the original denoising or flow-matching
objective and $\lambda_{\mathrm{RecCAR}}\geq 0$ controls the regularization
strength. For parameter-efficient adaptation, we keep the pretrained backbone
frozen and optimize LoRA parameters on its attention projections. The exact
adapted layers are backbone-specific and are detailed in the corresponding
experimental sections.

\method{} requires no additional correspondence supervision or auxiliary
model. It uses cross-modal structure already present in the pretrained joint
generator to strengthen the reciprocal pathway back into video. The same
formulation is applied to both video--motion and video--audio generation.

\begin{figure*}
    \centering
    \includegraphics[width=0.9\linewidth]{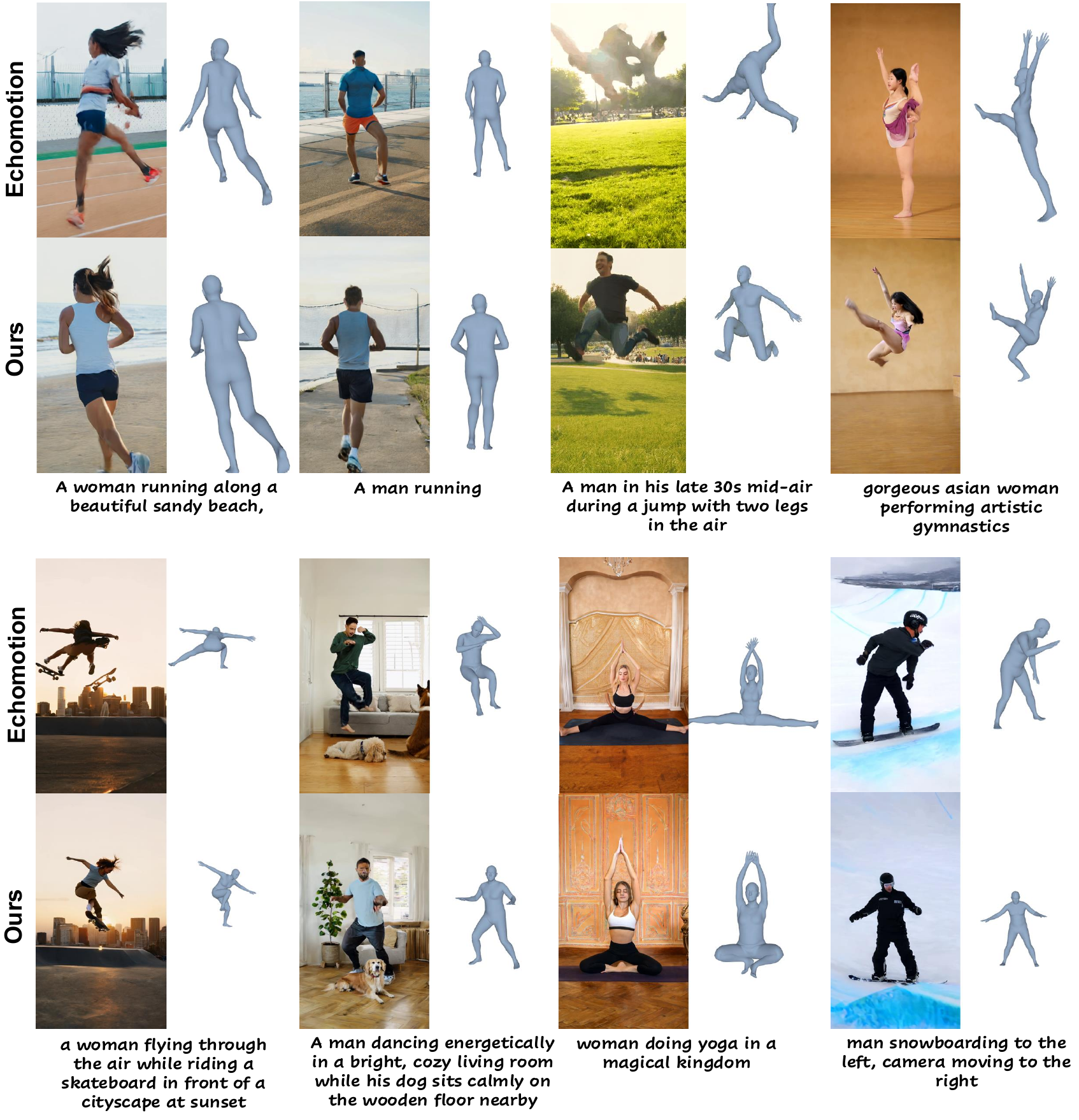}
    \caption{Qualitative comparison of motion reconstruction across six diverse scenes (running, skateboarding, yoga, and casual indoor activity). For each scene, we show two representative frames alongside the corresponding generated motion. Top: results from EchoMotion. Bottom: results from \ourmethod{}. Our approach produces more accurate and temporally consistent pose estimates, better capturing fine-grained limb articulation and body orientation even under challenging conditions.}
\end{figure*}

\begin{table*}[t]
    \centering
    \caption{ Generating video and motion. Metrics from VBench are computed on the test set. EchoMotion + \ourmethod{} (ours) outperforms EchoMotion \cite{yang2026echomotion}, CoMoVi \cite{zhao2026comovi}, and FlowMo \cite{shaulov2025flowmo} across most metrics, with the largest gains observed in Human Anatomy.}
    \label{tab:vm-test1000}
    \resizebox{\textwidth}{!}{%
    \begin{tabular}{lccccccccc}
        \toprule
        \textbf{Method} & 
        \textbf{Human} & 
        \textbf{Motion}&
        \textbf{Dynamic}&
        \textbf{Aesthetic}& 
        \textbf{Imaging} & 
        \textbf{Temporal} & 
        \textbf{Appear-} & 
        \textbf{Background}& 
        \textbf{Subject} \\
        \textbf{} & 
        \textbf{Anatomy} & 
        \textbf{Smooth-}&
        \textbf{Degree} & 
        \textbf{Quality}& 
        \textbf{Quality} & 
        \textbf{Flickering} &
        \textbf{ance} &
        \textbf{Consis-} & 
        \textbf{Consis-} \\
        \textbf{} & 
        \textbf{} & 
        \textbf{ness}&
        \textbf{} & 
        \textbf{}& 
        \textbf{} & 
        \textbf{} &
        \textbf{Style} &
        \textbf{tency} & 
        \textbf{tency} \\
        \midrule
        CoMoVi & $0.67$ & $0.99$ & $0.43$ & $0.46$ & $0.6846$ & $0.9802$ & $0.2184$ & $0.9208$ & $0.9063$ \\
        FlowMo & $0.65$ & $0.99$ & $0.62$ & $0.52$ & $0.5908$ & $\mathbf{0.9883}$ & $0.2647$ & $0.9339$ & $0.8941$ \\

        \midrule
        EchoMotion & $0.69$ & $0.98$ & $0.83$ & $0.56$ & $0.6785$ & $0.9643$ & $0.2814$ & $0.9346$ & $0.9108$ \\
        EchoMotion + \ourmethod{} (ours) & $\mathbf{0.75}$ & $0.99$ & $\mathbf{0.83}$ & $\mathbf{0.57}$ & $\mathbf{0.6869}$ & $0.9734$ & $\mathbf{0.2841}$ & $\mathbf{0.9406}$ & $\mathbf{0.9245}$ \\
        \bottomrule
    \end{tabular}%
    }
\end{table*}

\begin{table*}[t]
    \centering
    \small
    \caption{Results on T2AV-compass (500 prompts). Best value per column is in \textbf{bold}, arrows ($\uparrow,\downarrow$) indicate the direction of improvement. \textbf{Metrics}: \textit{AV Desync} (Abs./Pred. temporal offset), \textit{Audio Realism} (AAS/MTC perceptual scores), \textit{Video Aesth.} (SigLIP visual score), \textit{Video Quality} (Overall, Aesthetic, and Technical video fidelity), \textit{AudioBox Aesthetics} (PQ: Production Quality, CU: Content Usability, CE: Content Expressiveness, PC: Production Complexity), \textit{Speech} (NISQA MOS predictor), and \textit{Alignment} (T-V: Text-Video, T-A: Text-Audio, A-V: audio--video cross-modal cosine similarity).}
    \label{tab:t2av_transposed}
    \setlength{\tabcolsep}{3.5pt}
    \resizebox{\linewidth}{!}{%
    \begin{tabular}{l cc cc c ccc}
        \toprule
        
        & \multicolumn{2}{c}{\textbf{AV Desync} $\downarrow$}
        & \multicolumn{2}{c}{\textbf{Audio Realism} $\uparrow$}
        & \textbf{Video Aesth.} $\uparrow$
        & \multicolumn{3}{c}{\textbf{Video Quality} $\uparrow$} \\
        
        \cmidrule(lr){2-3}
        \cmidrule(lr){4-5}
        \cmidrule(lr){6-6}
        \cmidrule(lr){7-9}
        
        \textbf{Method}
        & \textbf{Abs.} & \textbf{Pred.}
        & \textbf{AAS} & \textbf{MTC}
        & \textbf{SigLIP}
        & \textbf{Overall} & \textbf{Aesth.} & \textbf{Tech.} \\
        
        \midrule

        \textbf{UniAVGen}
        & $0.866 \pm .031$ & $0.088 \pm .031$
        & $4.054 \pm 0.038$ & $3.763 \pm .095$
        & $4.510 \pm .003$
        & $\mathbf{0.809 \pm .003}$ & $\mathbf{0.9964 \pm .0002}$ & $0.119 \pm .001$ \\
        
        \midrule

        \textbf{JavisDiT++}
        & $1.078 \pm .027$ & $0.307 \pm .054$
        & $3.210 \pm .038$ & $2.845 \pm .041$
        & $\mathbf{4.752 \pm .035}$
        & $0.71 \pm .006$ & $0.9940 \pm .0010$ & $0.097 \pm .001$ \\

        \textbf{ITS-JavisDiT++}
        & $1.040 \pm .030$ & $0.250 \pm .043$
        & $3.045 \pm .021$ & $2.856 \pm .067$
        & $4.651 \pm .035$
        & $0.72 \pm .006$ & $0.9941 \pm .0010$ & $0.097 \pm .001$ \\

        \textbf{JavisDiT++ + \ourmethod{} (ours)} 
        & $1.046 \pm .028$ & $0.238 \pm .057$ & $3.383 \pm .148$ & $2.900 \pm .045$ & $4.620 \pm .034$ & $0.74 \pm .006$ & $0.9941 \pm .0010$ & $0.102 \pm .001$ \\
        \midrule
        \textbf{LTX-2}
        & $0.804 \pm .023$ & $0.095 \pm .048$
        & $4.180 \pm .036$ & $3.920 \pm .040$
        & $4.600 \pm .027$
        & $0.72 \pm .006$ & $0.9942 \pm .0005$ & $0.115 \pm .001$ \\

        \textbf{ITS-LTX-2}
        & $0.792 \pm .034$ & $0.075 \pm .048$
        & $4.202 \pm .028$ & $3.934 \pm .036$
        & $4.625 \pm .029$
        & $0.68 \pm .005$ & $0.9944 \pm .0001$ & $0.106 \pm .001$ \\
        
        \textbf{LTX-2 + \ourmethod{} (ours)}
        & $\mathbf{0.752 \pm .033}$ & $\mathbf{0.068 \pm .018}$
        & $\mathbf{4.225 \pm .035}$ & $\mathbf{3.975 \pm .039}$
        & $4.628 \pm .027$
        & $0.72 \pm .006$ & $0.9955 \pm .0005$ & $\mathbf{0.124 \pm .001}$ \\
        
        \bottomrule
    \end{tabular}%
    }

\vspace{0.8em}
    \resizebox{\linewidth}{!}{%
    \begin{tabular}{l cccc c ccc}
        \toprule
        
        & \multicolumn{4}{c}{\textbf{Audio Aesthetics} $\uparrow$}
        & \textbf{Speech} $\uparrow$
        & \multicolumn{3}{c}{\textbf{Alignment} $\uparrow$} \\
        
        \cmidrule(lr){2-5}
        \cmidrule(lr){6-6}
        \cmidrule(lr){7-9}
        
        \textbf{Method}
        & \textbf{PQ} & \textbf{CU} & \textbf{CE} & \textbf{PC}
        & \textbf{NISQA}
        & \textbf{T-V} & \textbf{T-A} & \textbf{A-V} \\
        
        \midrule

        \textbf{UniAVGen}
        & $6.402 \pm .019$ & $5.035 \pm .023$
        & $3.530 \pm .012$ & $2.368 \pm .011$
        & $1.268 \pm .009$
        & $0.0307 \pm .0022$ & $0.1044 \pm .0033$
        & $0.1690 \pm .0022$ \\

        \midrule

        \textbf{JavisDiT++}
        & $6.047 \pm .033$ & $5.707 \pm .052$
        & $2.780 \pm .014$ & $2.608 \pm .025$
        & $0.999 \pm .009$
        & $0.0242 \pm .0044$ & $0.0872 \pm .0030$ & $0.1318 \pm .0029$ \\

        \textbf{ITS-JavisDiT++}
        & $5.964 \pm .029$ & $5.243 \pm .044$
        & $2.773 \pm .013$ & $2.615 \pm .020$
        & $0.998 \pm .008$
        & $0.0241 \pm .0026$ & $0.8206 \pm .0030$ & $0.1324 \pm .0040$ \\

        \textbf{JavisDiT++ + \ourmethod{} (ours)} 
        & $6.334 \pm .026$ & $5.343 \pm .036$ & $2.707 \pm .026$ & $2.445 \pm .017$ & $0.936 \pm .012$ & $0.0277 \pm .0044$ & $0.0996 \pm .0041$ & $0.1613 \pm .0042$ \\
        \midrule
        \textbf{LTX-2}
        & $6.820 \pm .039$ & $6.483 \pm .053$
        & $4.684 \pm .063$ & $3.405 \pm .060$
        & $1.689 \pm .057$
        & $0.0344 \pm .0045$ & $0.1595 \pm .0041$
        & $\mathbf{0.2274 \pm .0046}$ \\

        \textbf{ITS-LTX-2}
        & $6.825 \pm .034$ & $\mathbf{6.519 \pm .033}$
        & $4.727 \pm .067$ & $3.463 \pm .054$
        & $1.584 \pm .037$
        & $0.0329 \pm .0074$ & $0.1615 \pm .0056$
        & $0.2134 \pm .0045$ \\
        
        \textbf{LTX-2 + \ourmethod{} (ours)}
        & $\mathbf{6.830 \pm .039}$ & $6.510 \pm .054$
        & $\mathbf{4.739 \pm .065}$ & $\mathbf{3.479 \pm .061}$
        & $\mathbf{1.695 \pm .057}$
        & $\mathbf{0.0353 \pm .0045}$ & $\mathbf{0.1619 \pm .0042}$
        & $0.2273 \pm .0047$ \\
        
        \bottomrule
    \end{tabular}%
    }
\end{table*}

\ignore{
\begin{figure*}[ht!]
    \centering
    \includegraphics[width=\linewidth]{figures/anatomy_vs_nmpjpe_gt_scatter_LLM.png}
    \caption{Caption}
    \label{fig:placeholder}
\end{figure*}
}

\section{Generating Video and Human Motion}
\label{sec:vm}

In this section, we apply our method to the problem of generating video together with generating a 3D model of human motion. 
Here, we build on top of EcoMotion~\cite{yang2026echomotion}. We now describe how we collect training data to fine-tune the model with \ourmethod{} in Section \ref{sec:vm-data}, then the training and evaluation procedures, and finally the results in Section \ref{sec:vm-results}. 

\subsection{Training data and procedure} \label{sec:vm-data}
We now describe how we curate high-quality paired data for fine-tuning \ourmethod{}. 
The challenge is that we need to collect data from generated videos, but they must also have structurally integral human anatomy and a faithful match to a 3D skeleton. 
We first selected 1,500 human text prompts from VidProM~\cite{wang2024vidprom} and an additional 1,000 prompts held out for testing. For each training prompt, we generated 16 video--motion pairs by varying the initialization random seeds. 
and considered the ~18000  who completed first as candidates for matching pairs. 

We develop a method to automatically quantify which pairs are good matches. To this end, we ran an experiment on a subset of 25 videos. 
For each of those videos, we estimated the 2D pose and compared it with the generated 3D pose rendered with the camera positioned at (0,0,0). 
We computed the mean per-joint position error (MPJPE) metric~\cite{ionescu2013human36m} and its normalized version NMPJPE~\cite{bogo2016keep}. 
We then sent them to LLM evaluation to measure which of these 25 videos pass the bar for being considered to be a good match. We find that video--motion pairs with MPJPE and NMPJPE scores below $0.15$ are good matches, and we use this threshold to filter the training set. 
The details of this experiment are given in the supplemental material . 
The filtering criterion allowed us to select those videos whose anatomy had structural integrity. To extract the 3D pose for training, we further estimated the 3D pose in the video using monocular 3D pose estimation,  CameraHMR~\cite{patel2025camerahmr}, which recovers camera-space SMPL motion sequences.
At the end of this process, we were left with  
4,292 video--motion training pairs. 
The dataset\footnote{\href{https://huggingface.co/datasets/ohad204/human-motion}{\texttt{huggingface.co/datasets/ohad204/human-motion}}} and the code\footnote{\href{https://github.com/ohad204/RecCAR/}{\texttt{github.com/ohad204/RecCAR}}} for \ourmethod{} are publicly available.

\paragraph{Implementation Details:}
We fine-tune EchoMotion by applying LoRA with rank 128 to all joint self-attention weights, following Eq (\ref{eq:m2v_logits}) and the loss in \eqref{eq:directional_correspondence_gap}. The model was fine-tuned for 10 epochs using \ourmethod{} with $\lambda_{\ourmethod{}}=0.01$.

\subsection{Evaluation Setup}
Since EchoMotion's original evaluation prompts were not released, we construct a test benchmark of 1,000 diverse prompts sampled from VidProM~\cite{wang2024vidprom} (not overlapping the training set). We use it to compare our method against EchoMotion~\cite{yang2026echomotion}, CoMoVi~\cite{zhao2026comovi}, and FlowMo~\cite{shaulov2025flowmo}.

\subsection{Results} \label{sec:vm-results}
We evaluate the generated videos using VBench~\cite{huang2023vbench, zheng2025vbench2}, a comprehensive benchmark for assessing video generation quality, specifically human anatomy, motion smoothness, dynamic degree, and aesthetic quality. Tables~\ref{tab:vm-test1000} compares our method \ourmethod{} against EchoMotion~\cite{yang2026echomotion}, CoMoVi~\cite{zhao2026comovi}, and FlowMo~\cite{shaulov2025flowmo} on 1000-prompt text-only test sets. 
Comparing to the EchoMotion baseline, adding \ourmethod{} significantly improves the metric of human anatomy and further improves all other metrics. 
The most pronounced improvement is observed in human anatomy, which is particularly relevant to the quality of the jointly generated motion and video, suggesting that aligning the reciprocal correspondence leads to generated video that is more consistent with the companion motion.
Notably, our method maintains a dynamic degree comparable to EchoMotion while substantially improving anatomical correctness, demonstrating that the gains arise from more accurate and coherent motion rather than simply generating slower or less challenging motions.

\begin{figure}
    \centering
    \includegraphics[width=\linewidth]{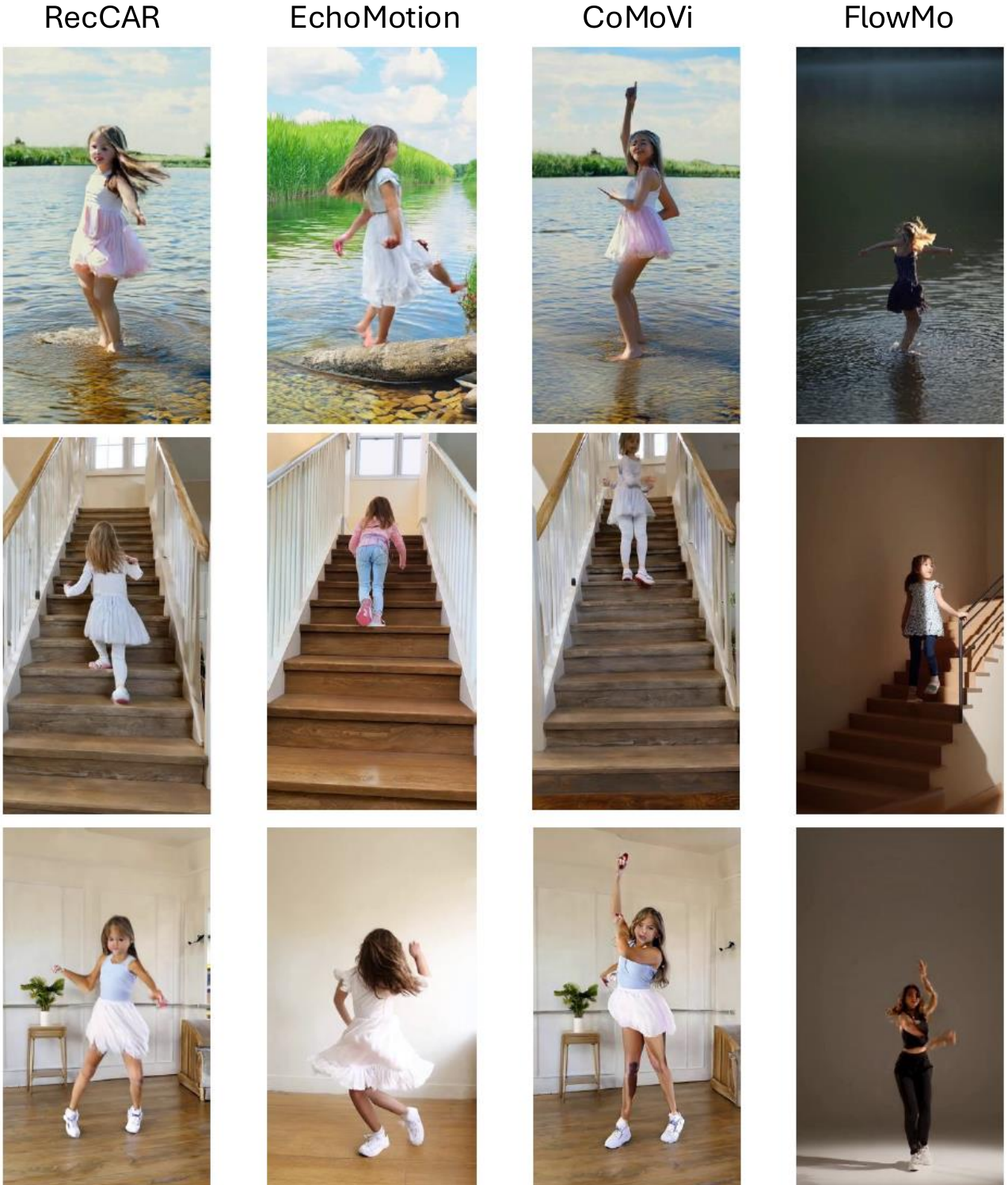}
    \caption{Qualitative comparison of \ourmethod{} against Echomotion, CoMoVi, and FlowMo across three representative scenes. \ourmethod{} produces frames with more natural and accurate human anatomy. At the same time, baseline methods exhibit varying degrees of extra limbs or unrealistic poses, such as legs pointing forward while the head faces backward. 
    }
\end{figure}

\section{Generating Video and Audio}
In this section, we further apply our approach to a different multi-modal generation task:  audio--video generation.
Here, we build on top of the backbones and models of LTX2~\cite{hacohen2026ltx2} and JavisDiT++~\cite{liu2026javisdit}.

\subsection{Training data} \label{sec:av-data}

For the audio--video track, we train on a randomly sampled subset of the VGGSound training set~\cite{chen2020vggsound}. VGGSound is a large-scale collection of audio--video clips spanning hundreds of everyday sound classes and recorded under diverse, real-world conditions. We use a subset of the dataset to keep training computationally efficient while retaining the diversity of its audio-visual content.

\paragraph{Implementation Details:}
We randomly sample $\sim$4,300 clips from the VGGSound training split and use them as provided, without additional filtering or preprocessing. We fine-tune the backbone using LoRA with rank 128 applied to all cross-attention weights, while keeping the remaining model parameters frozen. Training is performed for 10 epochs using our \ourmethod{} loss (Eq.~\ref{eq:directional_correspondence_gap}), with a loss weight of 0.01, matching the setting used for the video--motion track.

\begin{table}[t]
    \centering
    \caption{Comparison of audio--video synchronization and semantic alignment on AVGen-bench \cite{zhou2026avgen}. 
    DeSync is lower-is-better, while CLAP and AV-CLIP are higher-is-better.}
    \label{tab:av_comparison}
    \setlength{\tabcolsep}{3pt} 
    \resizebox{\linewidth}{!}{%
    \begin{tabular}{lccc}
        \toprule
        \textbf{Method} & \textbf{AV Desync } $\downarrow$ & \textbf{T-A (CLAP)} $\uparrow$ & \textbf{A-V (AV-CLIP)} $\uparrow$ \\
        
        \midrule
        UniAVGen
            & $0.656 \pm .07$ & $0.20 \pm .07$ & $0.76 \pm .04$ \\
  
        \midrule
        JavisDiT++ 
            & $0.518 \pm .01$ & $0.21 \pm .02$ & $0.72 \pm .06$ \\
        ITS-JavisDiT++
            & $0.507 \pm .05$ & $0.21 \pm .02$ & $\mathbf{0.76 \pm .05}$ \\
        \textbf{JavisDiT++ + \ourmethod{} (ours)}
            & $\mathbf{0.438 \pm .08}$ & $\mathbf{0.23 \pm .01}$ & $0.74 \pm .05$ \\   \midrule            
        LTX-2~\cite{hacohen2026ltx2} 
            & $0.424 \pm .05$ & $0.22 \pm .01$ & $0.74 \pm .05$ \\            
        ITS-LTX-2
            & $0.447 \pm .06$ & $0.21 \pm .02$ & $0.73 \pm .05$ \\ 
        \textbf{LTX-2 + \ourmethod{} (ours)}
            & $\mathbf{0.406 \pm .05}$ & $0.22 \pm .01$ & $0.74 \pm .05$ \\
        \bottomrule
    \end{tabular}%
    }
\end{table}

\subsection{Evaluation Setup}

We evaluate our approach on two audio--video generation benchmarks: 
T2AV-Compass~\cite{cao2026t2avcompass} and AVGen-Bench~\cite{zhou2026avgen}.
T2AV-Compass contains 500 challenging prompts and evaluates generated audio--video content across multiple dimensions of perceptual quality and cross-modal consistency.
AVGen-Bench contains 235 curated prompts, with particular emphasis on audio--video synchronization and semantic alignment.

We report the metrics provided by each benchmark.
For synchronization, we use \textbf{AV Desync}, which measures the temporal offset between the generated audio and video.
Semantic consistency is evaluated through text-audio (\textbf{T-A}), audio--video (\textbf{A-V}) and text-video (\textbf{T-V}) alignment.
We additionally report benchmark-specific measures of audio realism and production quality, video aesthetics and technical quality, and speech quality.

\paragraph{Baselines.}
We compare \ourmethod{} against LTX-2~\cite{hacohen2026ltx2}, 
JavisDiT++~\cite{liu2026javisdit}, 
UniAVGen~\cite{zhang2025uniavgen}, 
and ITS~\cite{jung2026inference} inferred using both LTX-2 and JavisDiT++ as backbones.
UniAVGen, ITS-LTX2, and ITS-Javis++ are particularly relevant comparisons, as all explicitly aim to improve synchronization between generated audio and video.

\subsection{Results}

\begin{figure}
    \centering
    \includegraphics[width=0.84\linewidth]{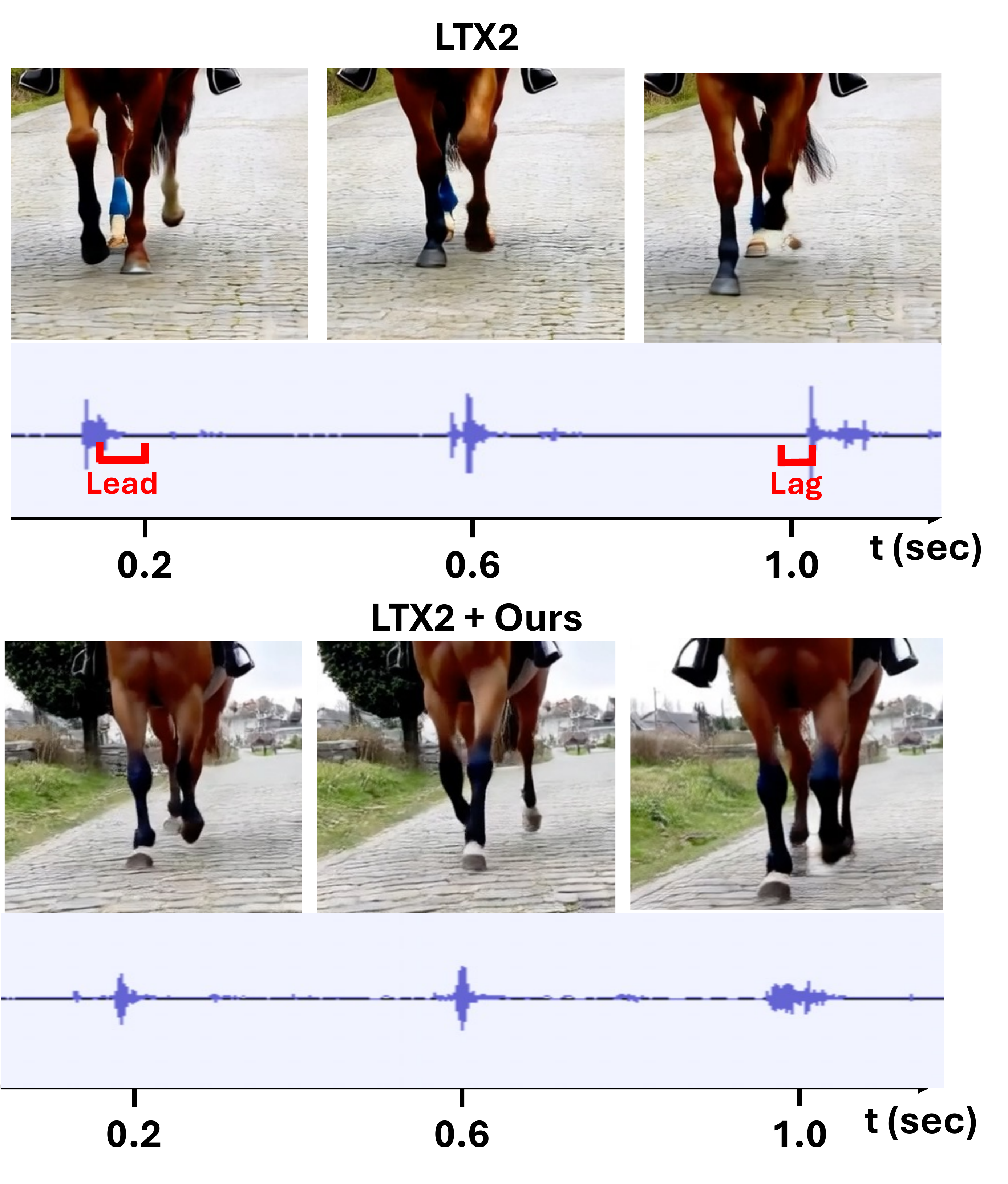}
    \caption{\textbf{Qualitative audio--video synchronization.} For a trotting horse, LTX-2 produces hoofbeat transients that lead or lag the hoof contacts and may exhibit visual artifacts such as missing hooves or multiple legs (top). \ourmethod{} better aligns the audio waveform with the horse's gait while preserving coherent anatomy (bottom).}
    \label{fig:audio-visual}
\end{figure}

Across both benchmarks, \ourmethod{} improves audio--video synchronization while preserving overall generation quality and semantic consistency. Figure~\ref{fig:audio-visual} illustrates this effect qualitatively: for a trotting horse, the baseline LTX-2 generates hoofbeat transients that lead or lag the visible hoof contacts, whereas \ourmethod{} better synchronizes the waveform with the horse's gait.

On T2AV-Compass (Table~\ref{tab:t2av_transposed}), applying \ourmethod{} to LTX-2 reduces both absolute and predicted AV Desync, achieving the best scores among all evaluated methods. The same objective also improves JavisDiT++ AV Desync metrics. For LTX-2, these synchronization gains coincide with improved audio realism (AAS/MTC), while video quality and audio–video alignment remain unchanged or improve slightly.

Table~\ref{tab:av_comparison} shows the same trend on AVGen-Bench. ITS yields only marginal DeSync reductions for both JavisDiT++ and LTX-2. In contrast, \ourmethod{} achieves substantially larger DeSync reductions for both backbones. CLAP and AV-CLIP are preserved or improved for both backbones, showing that the gains reflect better temporal coordination rather than a loss of semantic alignment.

\begin{table*}[t]
    \centering
    \caption{
    \textbf{Ablation on video--motion generation.}
    Comparison of baseline EchoMotion~\cite{yang2026echomotion}, standard fine-tuning, and \ourmethod{}. Standard fine-tuning degrades pose and anatomy metrics, whereas \ourmethod{} significantly improves them while maintaining smoothness, dynamic degree, and visual quality ($\downarrow$: lower is better; $\uparrow$: higher is better). We also ablate loss formulations and $\lambda_{RecCAR}$, showing that KL divergence and our selected learning rate yield optimal results.
    }
    \label{tab:motion_ablation}
        \setlength{\tabcolsep}{3pt} 
    \resizebox{\textwidth}{!}{%
    \begin{tabular}{clcccccc}
        \toprule
        & \textbf{Method} 
        & \textbf{N-MPJPE} 
        & \textbf{MPJPE} 
        & \textbf{Human} 
        & \textbf{Motion} 
        & \textbf{Dynamic} 
        & \textbf{Aesthetic} \\
        & 
        & $\downarrow$ 
        & $\downarrow$ 
        & \textbf{Anatomy $\uparrow$} 
        & \textbf{Smoothness $\uparrow$} 
        & \textbf{Degree $\uparrow$} 
        & \textbf{Quality $\uparrow$} \\
        \midrule

        \multirow{3}{*}{Ablation}
        & \textbf{EchoMotion + \ourmethod{}} (ours, $\lambda_{RecCAR}=0.01$)
            & $\mathbf{0.148 \pm 0.01}$
            & $\mathbf{0.29 \pm 0.04}$
            & $\mathbf{0.78 \pm 0.02}$
            & $\mathbf{0.99 \pm 0.001}$
            & $\mathbf{0.85 \pm 0.05}$
            & $\mathbf{0.57 \pm 0.01}$ \\
        
        & EchoMotion~\cite{yang2026echomotion}
            & $0.197 \pm 0.01$
            & $0.34 \pm 0.04$
            & $0.70 \pm 0.02$
            & $0.98 \pm 0.005$
            & $0.85 \pm 0.05$
            & $0.56 \pm 0.01$ \\

        & EchoMotion + standard fine-tuning
            & $0.208 \pm 0.01$
            & $0.35 \pm 0.04$
            & $0.66 \pm 0.04$
            & $0.97 \pm 0.001$
            & $0.84 \pm 0.05$
            & $0.57 \pm 0.01$ \\

        \midrule

        \multirow{3}{*}{\shortstack[c]{Sensitivity \\ to loss\\and $\lambda_{RecCAR}$:}}
        & EchoMotion + \ourmethod{} (cosine)
            & $0.165 \pm 0.02$
            & $0.31 \pm 0.05$
            & $0.71 \pm 0.03$
            & $0.99 \pm 0.001$
            & $0.85 \pm 0.05$
            & $0.56 \pm 0.01$ \\              

        & EchoMotion + \ourmethod{} ($\lambda_{RecCAR}=0.001$)
            & $0.198 \pm 0.02$
            & $0.32 \pm 0.05$
            & $0.65 \pm 0.04$
            & $0.98 \pm 0.001$
            & $0.84 \pm 0.05$
            & $0.56 \pm 0.01$ \\  

        & EchoMotion + \ourmethod{} ($\lambda_{RecCAR}=0.1$)
            & $0.322 \pm 0.02$
            & $0.78 \pm 0.06$
            & $0.61 \pm 0.03$
            & $0.95 \pm 0.01$
            & $0.76 \pm 0.05$
            & $0.41 \pm 0.01$ \\  
            
        \bottomrule
    \end{tabular}%
    }
\end{table*}

\begin{table}[t!]
    \centering
    \caption{
\textbf{Ablation on video--audio generation on AVGen-Bench~\cite{zhou2026avgen}.} Comparing pretrained, standard fine-tuned, and \ourmethod{} variants of JavisDiT++~\cite{liu2026javisdit} and LTX-2~\cite{hacohen2026ltx2}, standard fine-tuning barely improves synchronization. In contrast, \ourmethod{} achieves the lowest DeSync across both backbones while maintaining or boosting CLAP~\cite{elizalde2023clap} and AV-CLIP~\cite{iashin2024synchformer}.
    }
    \label{tab:audio_ablation}
    \setlength{\tabcolsep}{3pt} 
    \resizebox{\linewidth}{!}{%
    \begin{tabular}{lccc}
        \toprule
        \textbf{Method}
        & \textbf{DeSync [Sec]} $\downarrow$
        & \textbf{CLAP} $\uparrow$
        & \textbf{AV-CLIP} $\uparrow$ \\
        \midrule
        JavisDiT++ \cite{liu2026javisdit}
            & $0.518 \pm .01$ & $0.21 \pm .02$ & $0.72 \pm .06$ \\

        JavisDiT++ + standard fine-tuning
            & $0.498 \pm .08$ & $0.23 \pm .01$ & $0.74 \pm .05$ \\ 
            
        \textbf{JavisDiT++ + \ourmethod{} (ours})
            & $\mathbf{0.438 \pm .08}$ & $0.23 \pm .01$ & $0.74 \pm .05$ \\ 

        \midrule
        
        LTX-2~\cite{hacohen2026ltx2}
            & $0.424 \pm .05$
            & $0.22 \pm .01$
            & $0.74 \pm .05$ \\

        LTX-2 + standard fine-tuning
            & $0.421 \pm .04$
            & $0.22 \pm .01$
            & $0.74 \pm .05$ \\

        \textbf{LTX-2 + \ourmethod{} (ours)}
            & $\mathbf{0.406 \pm .05}$
            & $0.22 \pm .01$
            & $0.74 \pm .05$ \\
        \bottomrule
    \end{tabular}%
    }
\end{table}

\section{Ablation Study}
\label{sec:ablation}

We ablate \ourmethod{} separately on the two joint-generation settings to answer a central question: do improvements come simply from additional fine-tuning on the same training data, or specifically from our cross-attention alignment objective? In both settings, we compare the original backbone, standard fine-tuning on the same data without our loss, and the full \ourmethod{} model.

\paragraph{Video--Motion.}
Table~\ref{tab:motion_ablation} isolates the contribution of \ourmethod{} for joint video--motion generation, evaluated on VBench human-anatomy on a subset of 50 prompts~\cite{huang2023vbench,zheng2025vbench2}. Simply fine-tuning EchoMotion~\cite{yang2026echomotion} on the same curated training data does not improve video--motion consistency: N-MPJPE increases, MPJPE increases, and the Human Anatomy score decreases. Thus, additional training on our curated data alone cannot explain the gains.

In contrast, adding \ourmethod{} produces a clear improvement on all three metrics that directly reflect video--motion consistency. Relative to the original EchoMotion model, N-MPJPE decreases substantially, MPJPE decreases substantially, and Human Anatomy increases. Importantly, these improvements do not come from simplifying the generated motion: Dynamic Degree remains unchanged, while Motion Smoothness and Aesthetic Quality are preserved or slightly improved. 
These results show that the gains arise from the correspondence alignment objective of \ourmethod{}, rather than from generic fine-tuning on the same data.

\paragraph{Video--Audio.}
Table~\ref{tab:audio_ablation} shows the same controlled comparison for joint audio--video generation on  prompts from AVGen-Bench~\cite{zhou2026avgen}. Standard fine-tuning of LTX-2~\cite{hacohen2026ltx2} on the same training data has almost no effect on synchronization, with DeSync barely changing. In contrast, adding \ourmethod{} reduces DeSync relative to the original LTX-2 backbone.
At the same time, CLAP~\cite{elizalde2023clap} and AV-CLIP~\cite{iashin2024synchformer} remain unchanged. The synchronization improvement therefore does not come at the expense of semantic correspondence between modalities. Instead, \ourmethod{} selectively improves the temporal coordination between the jointly generated audio and video while preserving their semantic alignment. Together with the video--motion results, this shows that the same cross-attention alignment principle generalizes across two substantially different companion modalities.

\begin{figure}[t]
    \centering
    \includegraphics[width=0.83\linewidth]{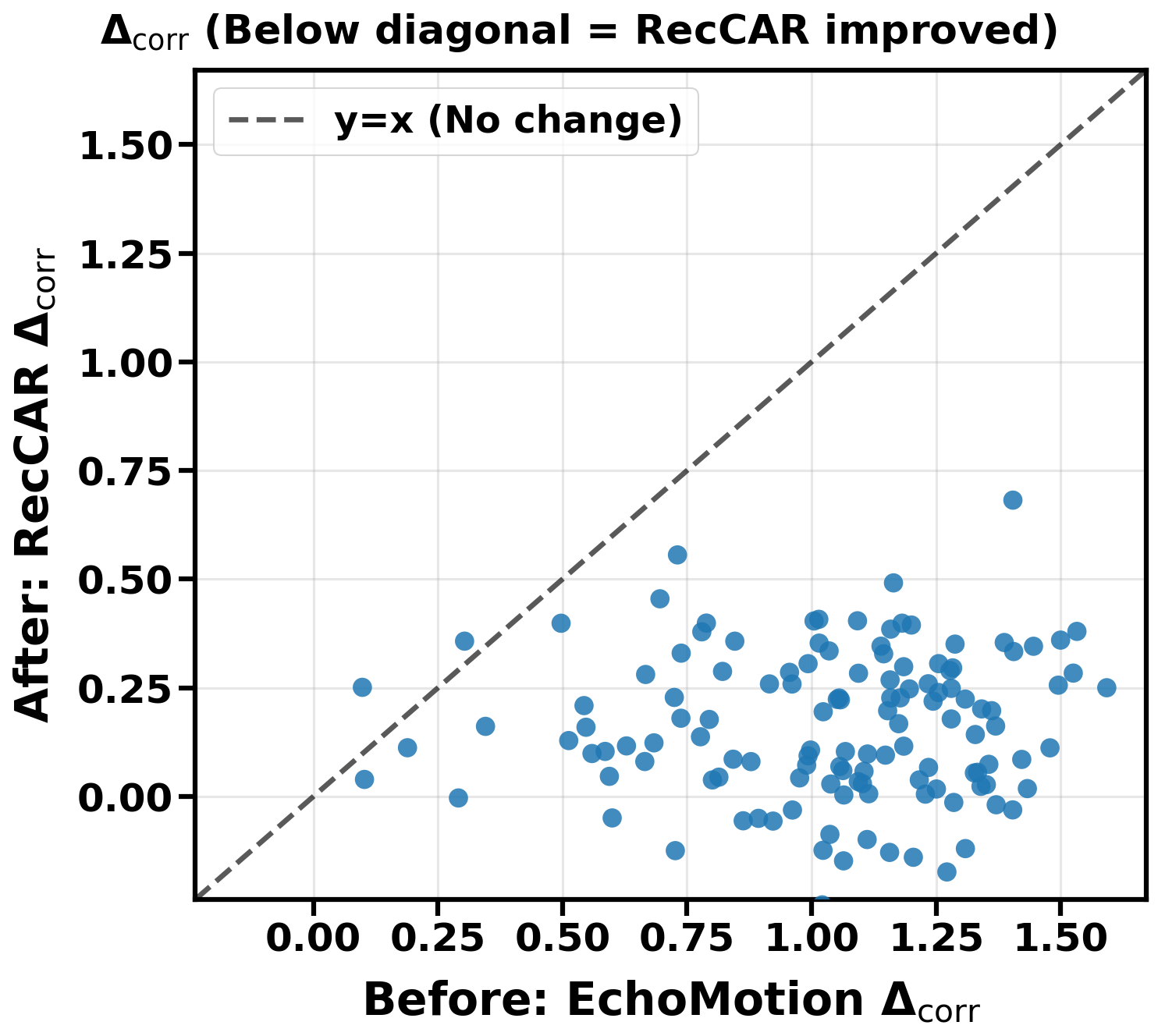}
    \caption{
    \textbf{Per-prompt reciprocal correspondence gap before vs. after \ourmethod{}.} Points below the $y=x$ line denote reduced $\Delta_{\text{corr}}$.
    }
    \label{fig:delta-corr}
\end{figure}

\clearpage
{
    \small
    \bibliographystyle{ieeenat_fullname}
    \bibliography{main}
}
\clearpage

\section{Appendix}

\subsection{Anatomy-Based Calibration of the Motion Filtering Criterion}
\label{sec:anatomy_calibration}

To identify video--motion pairs that provide reliable supervision for training, we calibrate a motion-fidelity threshold using an independent assessment of the anatomical quality of the humans appearing in the generated videos. The motivation is that low motion reconstruction error is not necessarily sufficient on its own: a generated video may have low error with respect to the target motion while still containing anatomically implausible humans. We therefore investigate whether motion fidelity, as measured by pose error, is associated with the visual plausibility of the generated human.

For this analysis, we randomly selected 25 generated videos from the full
generation set. For each video, we extracted the 2D human pose using
MediaPipe~\cite{lugaresi2019mediapipe}. We use the same MediaPipe skeleton and joint definition for all videos, ensuring a consistent correspondence between the estimated 2D pose and the joints of the associated generated 3D motion. Since the camera
parameters used to generate each motion are known, we projected the
corresponding 3D motion into the image coordinate system of the generated
video. We compared it with the 2D pose estimated from the RGB frames. We then computed the mean per-joint position error (MPJPE)~\cite{ionescu2013human36m}, averaged jointly over frames and joints.

To obtain an independent estimate of anatomical plausibility, we evaluated the same RGB videos using the Gemini LLM as an independent evaluator. Importantly, Gemini received only the generated RGB video and the corresponding text prompt and was not given the target 3D motion, the pose-estimation results, or any of the computed
motion errors. This prevents the model from directly inferring the anatomy
rating from the motion-fidelity measurements.

For each video, Gemini was asked to assess how faithful the depicted human is to real human anatomy, including whether the limbs correctly reflect human limbs and whether the body proportions are anatomically plausible. The model was instructed to provide an overall score between 1 and 10, with scores below 5 considered anatomically implausible. Gemini was also asked to provide a textual explanation for its assessment. Gemini evaluated the videos directly as full RGB videos rather than using individual frames, allowing the evaluator to consider anatomical consistency throughout the generated sequence.

\begin{figure}[b]
    \centering
    \includegraphics[width=0.83\linewidth]{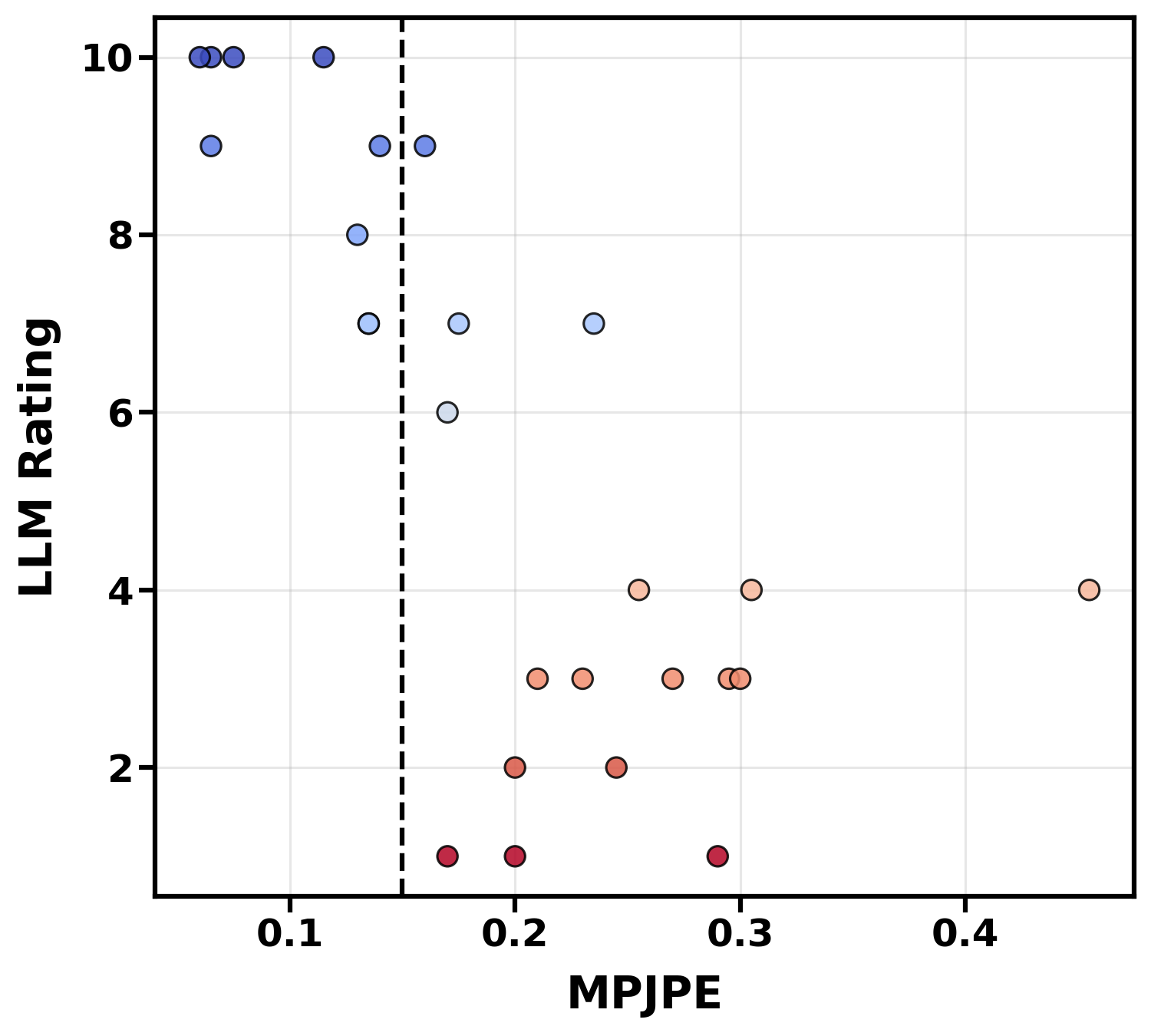}
    \caption{Calibration of the MPJPE filtering threshold. Gemini anatomy rating versus MPJPE for 25 generated videos; the dashed line marks the MPJPE $=0.15$ threshold.}
    \label{fig:anatomy_Vs_mpjpe}
\end{figure}

Figure~\ref{fig:anatomy_Vs_mpjpe} illustrates the relationship between the
motion error and the independent anatomy assessments. Each point corresponds to one of the 25 randomly selected videos, with the point color indicating the Gemini anatomical rating. A vertical dashed line marks the MPJPE~$=0.15$ threshold used for filtering the curated dataset. We observe a strong negative monotonic relationship between MPJPE and the Gemini rating. Thus, videos with lower motion error tend, on average, to be associated with more anatomically plausible generated humans.

Based on the observed relationship, we empirically set an MPJPE threshold of $0.15$, shown as the dashed vertical line in Figure~\ref{fig:anatomy_Vs_mpjpe}. This value separates a substantial portion of the samples with higher anatomical ratings from samples exhibiting larger motion errors and provides a simple criterion for identifying video--motion pairs whose generated humans are more likely to be anatomically faithful. The threshold is used only as a filtering criterion and is not intended to represent a universal anatomical error boundary.

We further require NMPJPE to be below the same threshold. While MPJPE captures the overall discrepancy between the generated video pose and the associated 3D motion, NMPJPE explicitly compensates for global scale differences. Jointly requiring
\[
\mathrm{MPJPE}<0.15
\qquad\text{and}\qquad
\mathrm{NMPJPE}<0.15
\]
therefore provides a more conservative criterion than using either metric
individually: a sample must exhibit both low absolute pose error and low
scale-normalized pose error. We apply this joint criterion when constructing the final set of reliable video--motion training pairs.

\subsection{Direct Validation of the Reciprocal Correspondence Gap}
\label{sec:delta_corr_validation}

Here we directly measure the quantity \ourmethod{} is designed to optimize: the reciprocal correspondence gap $\Delta_{\text{corr}}$ of Eq.~\eqref{eq:correspondence_gap} between the video-to-modality correspondence $C_{V \to M}$ and the modality-to-video correspondence $C_{M \to V}$, evaluated on held-out test prompts for the video--motion setting.

\paragraph{Attention mass allocation.}
Before turning to $\Delta_{\text{corr}}$, we examine a coarser,
complementary diagnostic: \emph{attention mass}, the fraction of a
query token's softmax weight placed on keys from the other
modality. This measures how strongly each direction is actually
used, independent of whether that attention is well-placed, meaning a
pathway with near-zero mass cannot meaningfully constrain the other
modality regardless of its accuracy, a question $\Delta_{\text{corr}}$
addressed separately below.

\begin{figure}[t]
    \centering
    \includegraphics[width=0.8\linewidth]{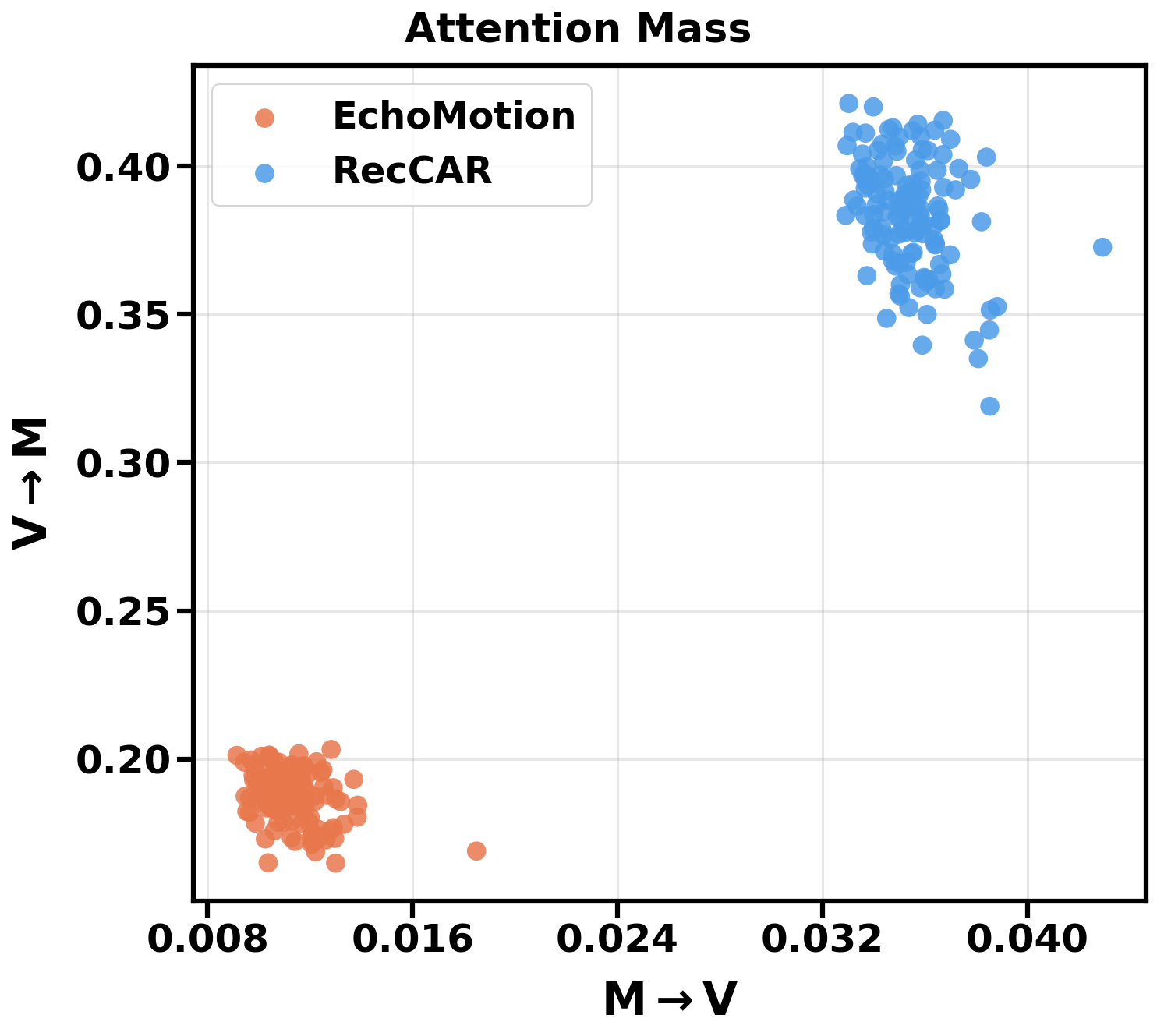}
    \caption{\textbf{MV vs. VM attention mass per test prompt},
    EchoMotion (orange) vs.\ \ourmethod{} (blue). The two clusters are cleanly separated, showing that \ourmethod{} consistently shifts cross-modal attention mass toward both the forward (V$\to$M) and, far more noticeably, the reciprocal (M$\to$V) direction across the test set.}
    \label{fig:attn-mass-scatter}
\end{figure}

For video-query tokens, we report the mass directed to motion tokens (motion$\to$video) against the mass retained on video tokens (video$\to$video); symmetrically, for motion-query tokens, we report the mass directed to video tokens (video$\to$motion) against the mass retained on motion tokens (motion$\to$motion). Table~\ref{tab:attn-mass} summarizes the direction of change per quadrant, and Figure~\ref{fig:attn-mass-scatter} shows the per-prompt distribution.
\ourmethod{} shifts attention mass toward the reciprocal
motion$\to$video direction, with a corresponding reduction in
video$\to$video self-attention. The forward video$\to$motion
direction shows the same qualitative pattern, drawing mass from
motion$\to$motion self-attention, though in relative terms the shift is
comparatively modest next to the change observed in the reciprocal
direction. Notably, in Figure~\ref{fig:attn-mass-scatter} the two
population clusters are cleanly separated with no overlap,
indicating this shift is consistent across the test set rather than
driven by a subset of prompts.

\begin{table}[t]
\centering
\caption{\textbf{Attention mass by cross-modal quadrant}, averaged
over test prompts. MV/VM denote the fraction of the query
modality's total attention mass directed to the other modality;
VV/MM denote the complementary self-attention mass.}
\label{tab:attn-mass}
\resizebox{0.47\textwidth}{!}{%
\begin{tabular}{lccc}
\toprule
Quadrant & EchoMotion & \ourmethod{} & $\Delta$ \\
\midrule
Video $\to$ Video   & 0.9887 & 0.9646 & $-2.4\%$ \\
Motion $\to$ Video  & 0.0113 & 0.0354 & $+213.3\%$ \\
Video $\to$ Motion  & 0.1871 & 0.3835 & $+104.9\%$ \\
Motion $\to$ Motion & 0.8130 & 0.6165 & $-24.2\%$ \\
\bottomrule
\end{tabular}%
}
\end{table}

\paragraph{Correspondence gap.}
We also directly compute $\Delta_{\text{corr}}$ per test prompt, before and after \ourmethod{} training, using the same cross-attention layers $\mathcal{R}$ and layer/head averaging.

Figure~\ref{fig:delta-corr} shows the per-prompt
$\Delta_{\text{corr}}$ EchoMotion versus \ourmethod{}
training: points below the diagonal indicate improved reciprocal
correspondence. The vast majority of test prompts fall below the
diagonal, showing that \ourmethod{} reduces the correspondence gap for
nearly all prompts rather than only on average.

\begin{figure}[t]
    \centering
    \includegraphics[width=0.8\linewidth]{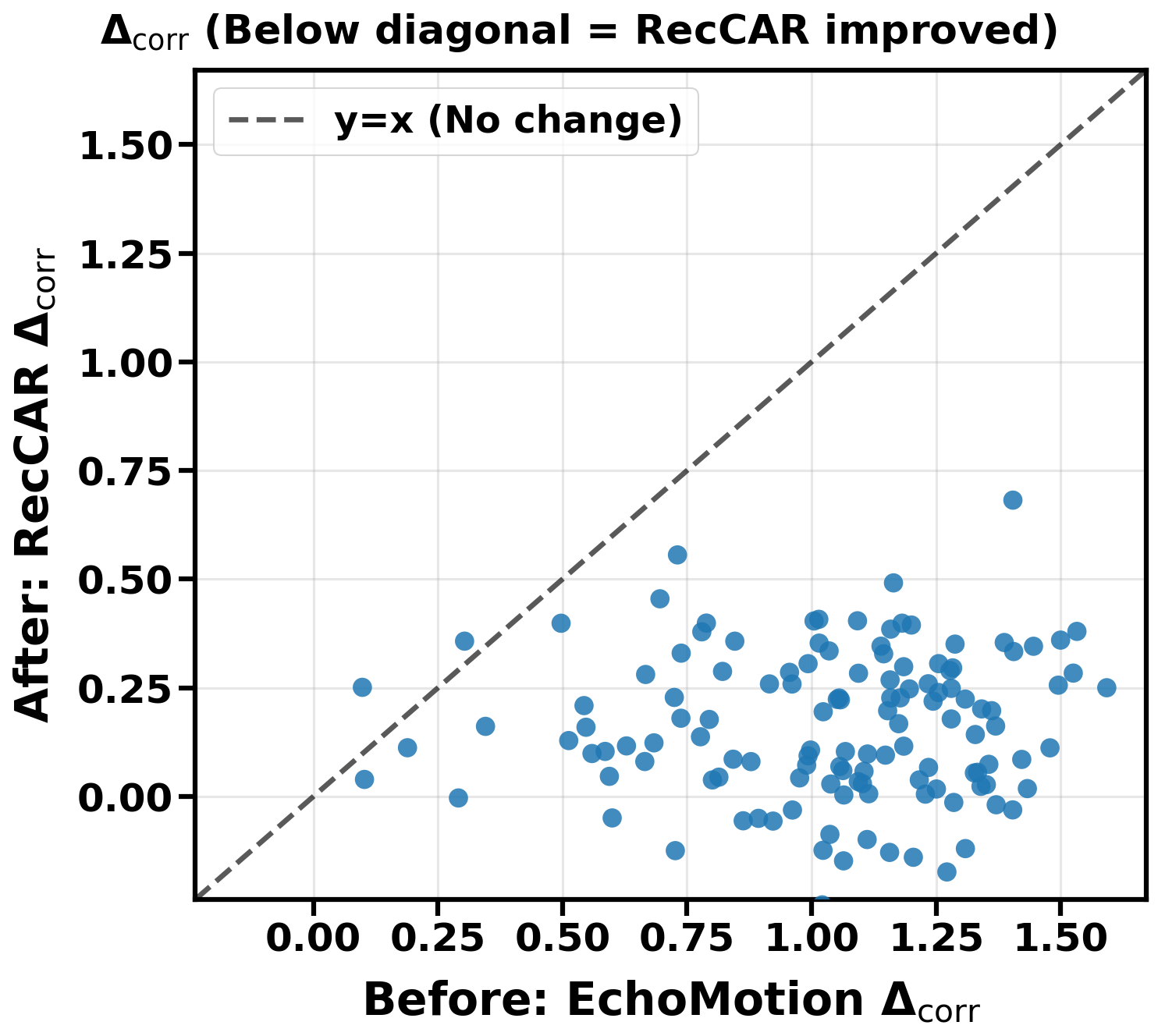}
    \caption{Per-prompt reciprocal correspondence gap before and after \ourmethod{}.** Each point compares $\Delta_{\text{corr}}$ before (x-axis) and after fine-tuning (y-axis). Points below $y=x$ indicate reduced correspondence gaps between $C_{V \to M}$ and $C_{M \to V}$. Most prompts fall below the diagonal, demonstrating consistent improvement across the test set.}
    \label{fig:delta-corr}
\end{figure}

Together, these results confirm that \ourmethod{}'s gains on downstream
metrics are accompanied by the intended mechanistic
change: the reciprocal modality-to-video pathway not only attends
more strongly to video (Table~\ref{tab:attn-mass}) but also comes to
agree more closely with the well-established video-to-modality
correspondence on \emph{where} in the video each token corresponds
(Figure~\ref{fig:delta-corr}), directly validating the
correspondence-gap hypothesis motivating \ourmethod{}.

\end{document}